\PassOptionsToPackage{table}{xcolor}

\pdfoutput=1
\documentclass[11pt]{article}

\usepackage[]{ACL2023}

\usepackage{soul}
\newcommand{\ctext}[3][RGB]{%
  \begingroup
  \definecolor{hlcolor}{#1}{#2}\sethlcolor{hlcolor}%
  \hl{#3}%
  \endgroup
}
\usepackage[table]{xcolor}

\usepackage{times}
\usepackage{latexsym}

\usepackage[T1]{fontenc}
\usepackage[utf8]{inputenc}

\usepackage{microtype}

\usepackage{inconsolata}

\usepackage{amsmath}
\DeclareMathOperator*{\argmax}{argmax} % thin space, limits underneath in displays
\usepackage{graphicx}
\usepackage{amsfonts}
\usepackage{color}
\definecolor{ForestGreen}{RGB}{34,139,34}
\usepackage{makecell}
\usepackage[margin=30pt]{subfig}
\usepackage{array}
\newcolumntype{P}[1]{>{\centering\arraybackslash}p{#1}}
\usepackage{arydshln}
\usepackage{multirow}

\DeclareUnicodeCharacter{307}{\c{c}}

\title{Balancing Lexical and Semantic Quality in Abstractive Summarization}

\author{Jeewoo Sul \and Yong Suk Choi\Thanks{ Corresponding author} \\
  Department of Computer Science \\
  Hanyang University, Seoul, Korea \\
  \texttt{\{jeewoo25, cys\}@hanyang.ac.kr} 
  }

\begin{document}
\maketitle
\begin{abstract}
An important problem of the sequence-to-sequence neural models widely used in abstractive summarization is \textit{exposure bias}. To alleviate this problem, re-ranking systems have been applied in recent years. Despite some performance improvements, this approach remains underexplored. Previous works have mostly specified the rank through the ROUGE score and aligned candidate summaries, but there can be quite a large gap between the lexical overlap metric and semantic similarity. In this paper, we propose a novel training method in which a re-ranker balances the lexical and semantic quality. We further newly define false positives in ranking and present a strategy to reduce their influence. Experiments on the CNN/DailyMail and XSum datasets show that our method can estimate the meaning of summaries without seriously degrading the lexical aspect. More specifically, it achieves an 89.67 BERTScore on the CNN/DailyMail dataset, reaching new state-of-the-art performance. Our code is publicly available at \url{https://github.com/jeewoo1025/BalSum}.
\end{abstract}

\section{Introduction}
The performance of sequence-to-sequence (Seq2Seq) neural models for abstractive summarization \cite{lewis-etal-2020-bart, nallapati-etal-2016-abstractive, see-etal-2017-get, zhang2020pegasus} has improved significantly. The dominant training paradigm of Seq2Seq models is that of Maximum Likelihood Estimation (MLE), maximizing the likelihood of each output given the gold history of target sequences during training. However, since the models generate the sequence in an auto-regressive manner at inference, the errors made in the previous steps accumulate in the next step thereby affecting the entire sequence. This phenomenon is known as \textit{exposure bias} \cite{NIPS2015_e995f98d, ICLR2016_Aurelio}. To mitigate this problem, re-ranking systems \cite{liu-etal-2021-refsum, liu-liu-2021-simcls, liu-etal-2022-brio, ravaut-etal-2022-summareranker} have recently been introduced to generate a more appropriate summary. 

\begin{figure}[!t]
\small{
    \centering
    \includegraphics[width=1.0\linewidth]{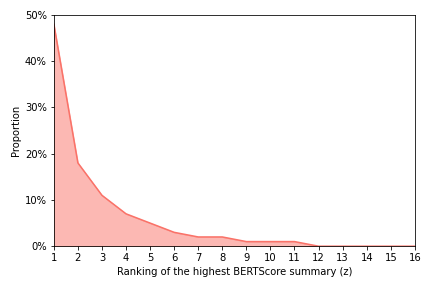}
    \caption{Distribution of $z$ (\%) for a base BART model on CNN/DM. Since a BART model generates a pool of 16 diverse beam search candidates, the X-axis ranges from 1 to 16. If $z=1$, it means that both ROUGE and BERTscore are high. As $z$ increases, the gap between ROUGE and BERTScore tends to increase. The Y-axis represents the proportion of $z$ in the test set. The distribution for XSum is in Appendix \ref{sec:appendix_xsum}.}
    \label{fig:1}
}
\end{figure}

There are two training objectives for applying re-ranking to abstractive summarization: \textit{contrastive learning} and \textit{multi-task learning}. The contrastive learning-based approaches deploy margin-based losses. SimCLS \cite{liu-liu-2021-simcls} and BRIO-Ctr \cite{liu-etal-2022-brio} train a large pre-trained model, such as RoBERTa \cite{2019RoBERTa} and BART \cite{lewis-etal-2020-bart}, to align the candidate summaries according to the quality. The authors use the ROUGE \cite{lin-2004-rouge} score as a quality measurement. The multi-task learning-based approaches combine at least two losses that perform different roles.  SummaReranker \cite{ravaut-etal-2022-summareranker} minimizes the average over the binary cross-entropy losses optimized for each evaluation metric. In addition, BRIO-Mul \cite{liu-etal-2022-brio} demonstrates that the combination of the contrastive and cross-entropy loss works complementarily and has better performance.  

In this paper, we analyze the three main drawbacks of existing re-ranking approaches. First, we argue that current methods focus excessively on ranking summaries in terms of lexical overlap. Inspired by \citet{zhong-etal-2020-extractive}, we conduct a preliminary study, by sorting candidate summaries in descending order based on the ROUGE score and then defining $z$ as the rank index of the highest BERTScore summary. As demonstrated in Fig. \ref{fig:1}, we can observe that there is a large gap between lexical overlap and semantic similarity. In a majority (52\%) of cases $z>1$. Second, despite more than half of the candidates with the same ROUGE score, previous studies do not accurately reflect quality measurements as they are trained with different ranks even if they have equal scores (Appendix \ref{sec:appendix_identical_score}). Lastly, for the first time, we find summaries with high lexical overlap but low semantic similarity as false positives (Appendix \ref{sec:appendix_example}). They can be noises during training phrase, which are not considered substantially in the prior works. 

To address these issues, we propose a novel training method in which a re-ranker balances lexical and semantic quality. Based on a two-stage framework, our model, named \textit{BalSum}, is trained on multi-task learning. We directly reflect the ROUGE score difference on a ranking loss to preserve the lexical quality as much as possible. Then, we use a contrastive loss with instance weighting to identify summaries whose meanings are close to the document. Specifically, we define novel false positives (semantic mistakes) and present a strategy to reduce their influence in ranking. Experiments on CNN/DM and XSum datasets demonstrate the effectiveness of our method. Notably, BalSum achieves an 89.67 BERTScore on CNN/DM, reaching a new state-of-the-art performance. 

\section{Method}
Our method follows the two-stage framework. Given a source document $D$, a function $\mathnormal{g}$ is to generate a pool of candidate summaries $\mathbb{C}=\{C_1, C_2, ..., C_m\}$ at the first stage:
\begin{equation}\label{eq1}
    \mathbb{C} \leftarrow g(D)
\end{equation}
Then, a function $\mathnormal{f}$ is to assign scores to each candidate and select the best summary $C^{*}$ with the highest score at the second stage: 
\begin{equation}\label{eq2}
    C^{*}=\argmax_{C_i \in \mathbb{C}} \{f(C_i,D)\}
\end{equation}
Our goal is to train the ranking model $f$ that identifies the correct summary from the outputs of the generation model $g$. 

\begin{figure}[t!]
    \centering
    \includegraphics[width=1.0\linewidth]{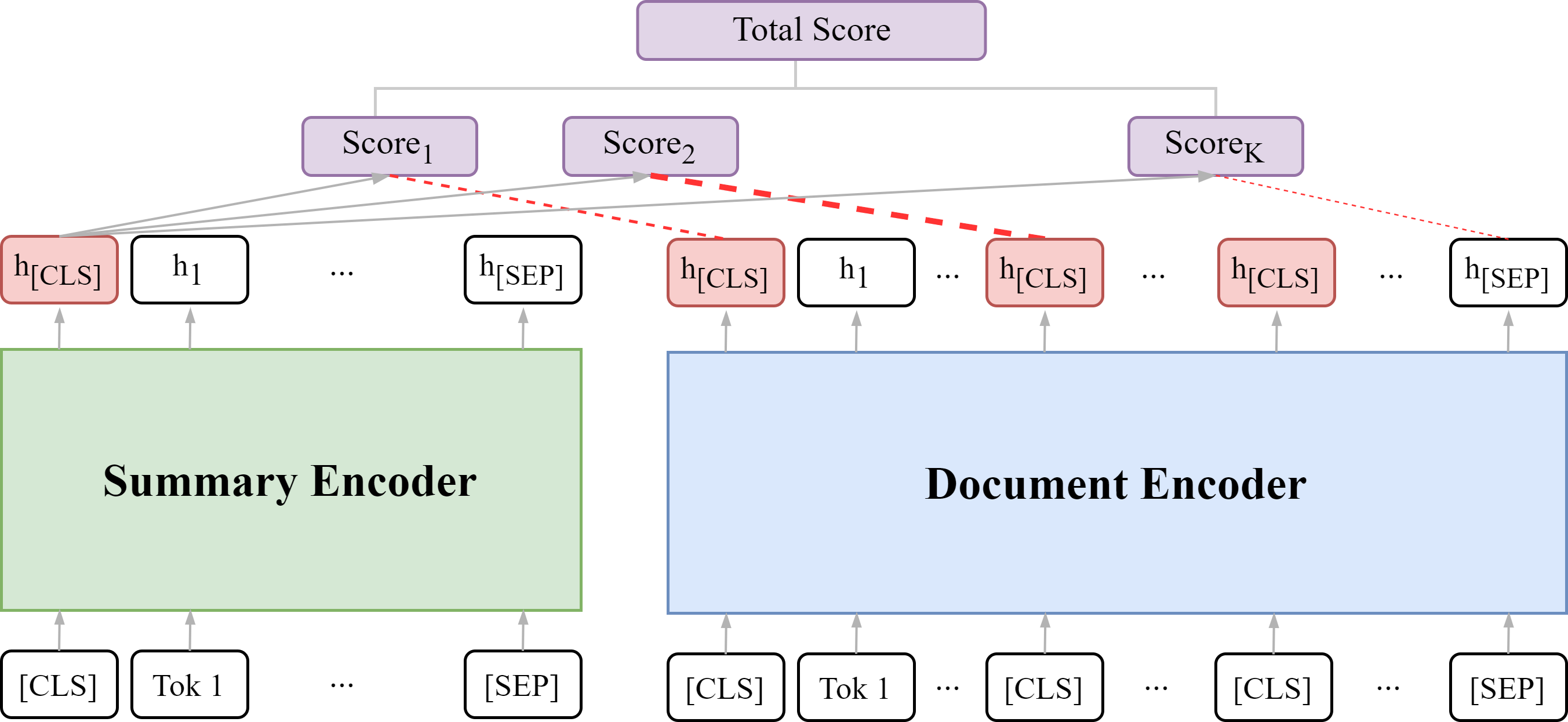}
    \caption{BalSum model architecture. The model predicts scores for candidate summaries based on the document. The thickness of the red dashed line indicates the magnitude of each score's weight. }
    \label{fig:2}
\end{figure}

\subsection{Model Architecture}
We start with a bi-encoder using RoBERTa-base \citep{2019RoBERTa} as a back-bone neural network. Inspired by \citet{sigir20_colbert}, we aim to capture rich semantic units at the sentence level. As shown in Fig. \ref{fig:2}, we insert the \textit{[CLS]} tokens in front of $K$ sentences in the document $D$ to let them encode into multi-vector representations. Then, we compute the individual score $Score_{k}$ which is modeled as an inner-product:  
\begin{equation}\label{eq3}
    Score_{k} = sim(E_1(C_i), E_k(D))
\end{equation}
where $E_1(C_i)$ and $E_k(D)(k=1,2,...,K)$ mean the representations of \textit{[CLS]} tokens for candidate summary $C_i$ and document $D$, respectively. We calculate the similarity score $f(C_i,D)$:
\begin{equation}\label{eq4}
\small{
\begin{aligned} 
    f(C_{i}, D) = \sum_{k=1}^{K}\frac{Score_{k}}{\sum_{j=1}^{K}Score_{j}}Score_{k} = \sum_{k=1}^{K}w_k \cdot Score_k   
\end{aligned}
}
\end{equation}
In Appendix \ref{sec:appendix_WeightAvg}, we show that our model can capture more information from documents at the sentence level.

\subsection{Training objective}
\begin{figure}[]
    \centering
    \includegraphics[width=1.0\linewidth]{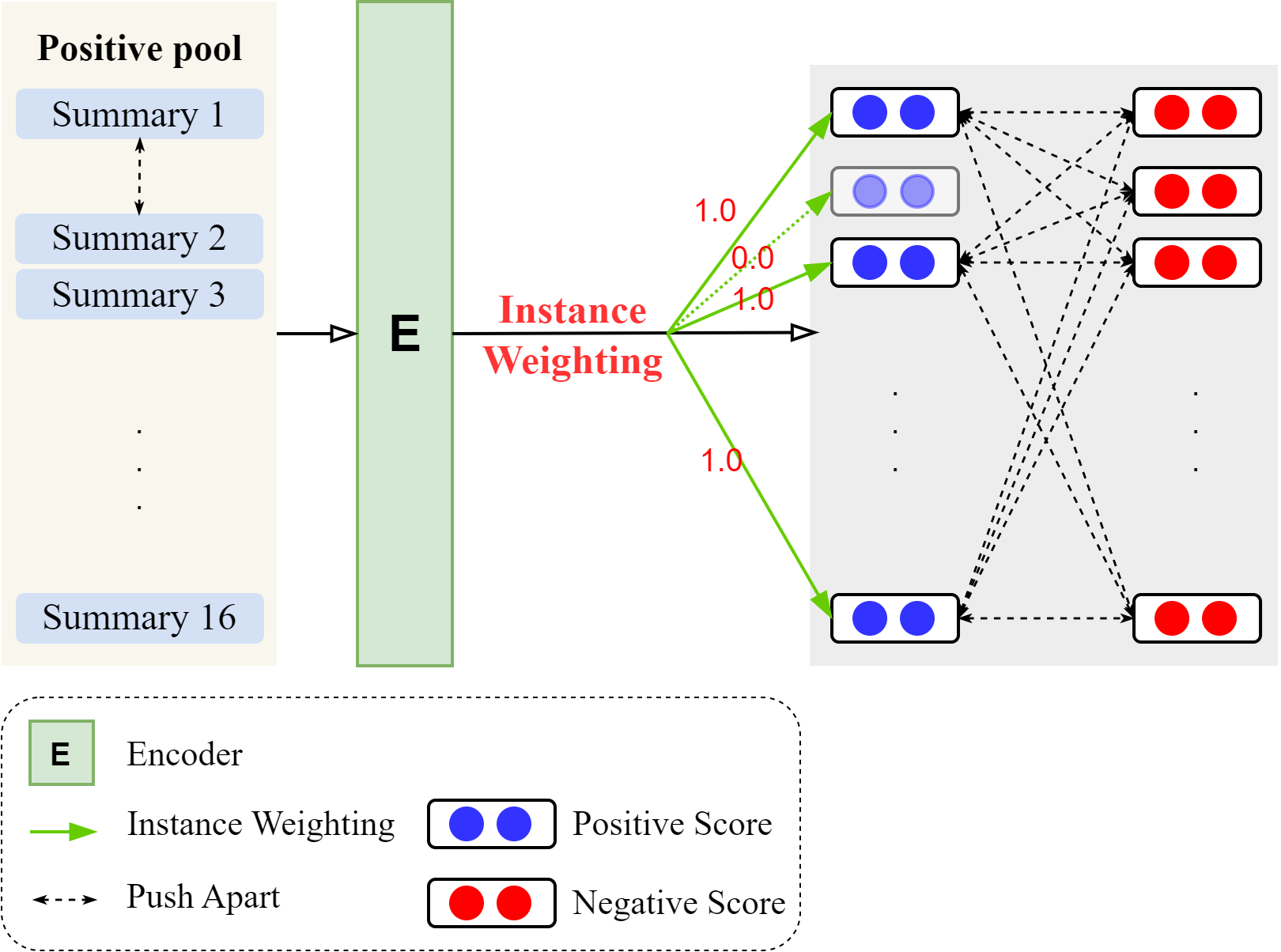}
    \caption{Overview of our proposed training objective.}
    \label{fig:3}
\end{figure}
\paragraph{Ranking Loss}
The core idea is that the higher the quality of the candidate summary, the closer to the document. We introduce a ranking loss to $f(\cdot)$:
\begin{equation}\label{eq5}
\small{
\begin{aligned}
    \mathcal{L}_{rank} = \sum_{i}\sum_{j>i} max(0, f(C_j,D)-f(C_i,D) \\
    \ + (-cost(C_i,S) + cost(C_j,S))\ast \lambda) \\
\end{aligned}
}
\end{equation}
where $S$ is the reference summary and $\lambda$ is the hyper-parameter.\footnote{We set $\lambda$ to 1.0 on CNN/DM and 0.1 on XSum.} Here, $cost(C_i,S)=1-M(C_i,S)$ is the margin, and $M$ is the automatic evaluation metric. We define it as ROUGE. We use the same metric in previous work \cite{liu-liu-2021-simcls, liu-etal-2022-brio}, but the difference is that our loss directly reflects the quality measure during training. In other words, the quality was not properly reflected before because different margin ($(j-i)*\lambda$) was assigned even if the candidate summaries had the same ROUGE score.

\paragraph{Contrastive Loss with Instance Weighting}
The construction of positive and negative pairs is the critical point in constrative learning. Therefore, we consider generated summaries from the same document as \textit{positive samples} and irrelevant summaries from other documents as \textit{negative samples}. Thus, we design a set of candidate summaries $\mathbb{C}$ in Eq. \ref{eq1} as \textit{positive} and a set of randomly sampled summaries $N$ as \textit{negative}.\footnote{As it is insensitive, we fix a negative strategy to random sampling in our experiments.} To identify summaries whose meanings are close to the document, we introduce a contrastive learning objective with instance weighting:
\begin{equation}\label{eq6}
    \mathcal{L}_{ctr} = \frac{1}{|\mathbb{C}|}\sum_{C_i \in \mathbb{C}}-log\frac{\alpha_{C_i} \times e^{f(C_i,D)}}{e^{f(C_i,D)}+\sum_{s_i \in N}e^{f(s_i,D)}}
\end{equation}
We newly define summaries that have a high lexical matching but a low semantic similarity as \textit{false positives}. Inspired by \citet{zhou-etal-2022-debiased}, we design an instance weighting method to reduce the influence of false positives. We produce the weights for positives using the SimCSE \citep{gao-etal-2021-simcse} which is the state-of-the-art model for the sentence representation task: 
\begin{equation}\label{eq7}
    \alpha_{C_i} = \begin{cases}
        0, & sim(C_i, S) < \phi \\
        1, & sim(C_i, S) \geq \phi
    \end{cases}
\end{equation}
where $\phi$ is a hyper-parameter of the instance weighting threshold, and $sim(\cdot)$ is the cosine similarity score evaluated by the SimCSE model. 

Finally, as shown in Fig. \ref{fig:3}, we combine the ranking (Eq. \ref{eq5}) and contrastive (Eq. \ref{eq6}) losses:
\begin{equation}\label{eq8}
    \mathcal{L} = \gamma_{1}\mathcal{L}_{rank} + \gamma_{2}\mathcal{L}_{ctr}
\end{equation}
where $\gamma$ is the scale factor of each loss and we find the optimal values ($\gamma_1=10, \gamma_2=0.1$) in Appendix \ref{sec:appendix_hyper}.

\section{Experiments}

\subsection{Datasets}
We experiment on two datasets, whose statistics are shown in Appendix \ref{sec:appendix_dataset}.

\textbf{CNN/DailyMail} \citep{NIPS2015_afdec700} is the most commonly used summarization dataset which contains articles from the CNN and DailyMail newspapers. 

\textbf{XSum} \citep{narayan-etal-2018-dont} is a one-sentence summary dataset from the British Broadcasting Corporation (BBC) for the years 2010 - 2017.

\subsection{Training Details}
We use diverse beam search \citep{VijayakumarCSSL16} to generate 16 candidate summaries. We start from pre-trained checkpoints of RoBERTa-base \citep{2019RoBERTa}. We train BalSum for five epochs. It takes 33 hours on CNN/DM and 22 hours on XSum on a single RTX 3090 GPU. More details are described in Appendix \ref{sec:appendix_details}.

\subsection{Main Results}
\begin{table}[t]
\small{
\begin{tabular}{lllll}
\Xhline{1pt} \\[-1.5ex]
\textbf{Model} &  \textbf{R-1} & \textbf{R-2} & \textbf{R-L} & \textbf{BS} \\[0.5ex]
\Xhline{1\arrayrulewidth} \\[-1.5ex]
BART*               & 44.16          & 21.28          & 40.90          & -               \\
BART$^\ddagger$    & 44.04          & 21.06          & 40.86          & 88.12          \\
Pegasus*            & 44.16          & 21.56          & 41.30          & -              \\[0.5ex]
\Xhline{1\arrayrulewidth}\\[-1.5ex]
BRIO-Mul*           & 47.78          & 23.55          & 44.57          & -              \\
BRIO-Mul$^\ddagger$ & \textbf{47.50} & \textbf{23.48} & 44.01 & 89.08          \\
BRIO-Ctr*           & 47.28          & 22.93          & 44.15          & -              \\
BRIO-Ctr$^\ddagger$ & 47.08          & 23.03          & \textbf{44.06} & 89.03          \\
SummaReranker* & 47.16          & 22.55          & 43.87          & 87.74          \\
SimCLS*       & 46.67          & 22.15          & 43.54          & -              \\
SimCLS$^\ddagger$ & 46.34 & 22.07 & 43.30 & 88.92      \\[0.5ex]
\Xhline{1\arrayrulewidth}\\[-1.5ex]
BalSum        & 46.58$^\dagger$ & 22.33$^\dagger$ & 43.49$^\dagger$ & \textbf{89.67$^\dagger$} \\[0.5ex]
\Xhline{1pt}\\[-1.5ex]
\end{tabular}
}
\vspace{-1\baselineskip}
\caption{\textbf{Results on CNN/DM}. R-1/2/L are the ROUGE-1/2/L $F_{1}$ scores. \textbf{BS} denotes BERTScore. *: results reported in the original papers. $\ddagger$: results from our own evaluation script. $\dagger$: significantly better than the baseline model (BART).}
    \label{tab:CNNDM}
\end{table}
In terms of the two-stage framework, we compare our results with SimCLS \citep{liu-liu-2021-simcls}, SummaReranker \citep{ravaut-etal-2022-summareranker}, and BRIO \citep{liu-etal-2022-brio}. We apply BalSum on top of each base model which is BART or PEGASUS. 

The results on CNN/DM are described in Table \ref{tab:CNNDM}. BalSum outperforms a base BART model, according to gains of 2.54/1.27/2.63 R-1/2/L. Notably, while it has comparable performances on ROUGE to previous models, it achieves an 89.67 BERTScore, reaching a new state-of-the-art performance. When ranking the candidate summaries, our model can estimate the meaning of summaries without seriously degrading the lexical aspect. We argue that this is because BalSum decreases more false positives than other ranking models. We provide fine-grained analyses for this result and present a case study in Sec.\ref{analysis}.

In addition, we apply our method on XSum, as shown in Table \ref{tab:XSum}. Though we use a different strategy to generate the validation and test data \footnote{We use 4 groups for diversity sampling, which results in 4 candidates. This is the same as SimCLS.}, our method improves a base PEGASUS with a small margin. We believe the one of reasons is that XSum is restricted to capturing diverse semantic units because it consists of much shorter summaries (one-sentence) than CNN/DM. 

\begin{table}[t]
\small{
\begin{tabular}{lllll}
\Xhline{1pt} \\[-1.5ex]
\textbf{Model} &  \textbf{R-1} & \textbf{R-2} & \textbf{R-L} & \textbf{BS} \\[0.5ex]
\Xhline{1\arrayrulewidth} \\[-1.5ex]
BART*               & 45.14          & 22.27          & 37.25          & -               \\
Pegasus*            & 47.21          & 24.56         & 39.25         & -              \\
Pegasus$^\ddagger$    & 46.82          & 24.44          & 39.07          & 91.93          \\[0.5ex]
\Xhline{1\arrayrulewidth}\\[-1.5ex]
BRIO-Mul*           & 49.07 & 25.59 & 40.40 & -              \\
BRIO-Mul$^\ddagger$ & \textbf{48.74} & \textbf{25.38} & \textbf{40.16} & \textbf{92.60} \\
BRIO-Ctr*           & 48.13 & 25.13 & 39.84 & -              \\
BRIO-Ctr$^\ddagger$ & 48.12 & 25.24 & 39.96 & 91.72          \\
SummaReranker* & 48.12 & 24.95 & 40.00 & 92.14 \\
SimCLS*       & 47.61 & 24.57 & 39.44  & -              \\
SimCLS$^\ddagger$ & 47.37 & 24.49 & 39.31 & 91.48 \\[0.5ex]
\Xhline{1\arrayrulewidth}\\[-1.5ex]
BalSum & 47.17$^\dagger$ & 24.23 & 39.09 & 91.48 \\[0.5ex]
\Xhline{1pt}\\
\end{tabular}
}
\vspace{-1\baselineskip}
\caption{\textbf{Results on XSum}. R-1/2/L are the ROUGE-1/2/L $F_{1}$ scores. \textbf{BS} denotes BERTScore. *: results reported in the original papers. $\ddagger$: results from our own evaluation script. $\dagger$: significantly better than the baseline model (PEGASUS).}
    \label{tab:XSum}
\end{table}

\subsection{Analysis} \label{analysis}
\paragraph{Weighting Threshold $\phi$} Intuitively, the larger the weighting threshold, the lower false positives.  We train our model with different instance weighting thresholds from $0.7$ to $0.9$. In Table \ref{tab:Abl_2}, the highest threshold ($\phi=0.9$) shows the best performance and it rises largely to $0.3$ BERTScore compared to when not applied. We also find that increasing the threshold leads to performance improvement. Therefore, we demonstrate that false positives can be considered noise in training.

\begin{table}[!t]
\small{
    \centering
    \begin{tabular}{lcccccc}
    \Xhline{1pt} \\[-1.5ex]
            $\phi$ & N/A   & 0.7   & 0.75  & 0.8   & 0.85  & 0.9   \\[0.5ex]
            \Xhline{1\arrayrulewidth} \\[-1.5ex]
            \textbf{BS} & 89.37 & 89.35 & 89.36 & 89.63 & 89.37 & \textbf{89.67} \\[0.5ex]
    \Xhline{1pt}\\
    \end{tabular}
    \vspace{-1\baselineskip}
    \caption{BERTScore (noted \textbf{BS}) results with different weighting threshold $\phi$ on CNN/DM. ``N/A'': no instance weighting.}
    \label{tab:Abl_2}
}
\end{table}

\begin{table}[!t]
    \centering
    \small{
    \resizebox{.5\textwidth}{!}{%
    \begin{tabular}{lllllll}
    \Xhline{1pt} \\[-1.5ex]
    \textbf{Model} & \textbf{BS@1} & \textbf{BS@3} & \textbf{BS@5} & \textbf{R@1} & \textbf{R@3} & \textbf{R@5} \\[0.5ex]
    \Xhline{1\arrayrulewidth} \\[-1.5ex]
        Oracle (R)     & 90.77         & 90.42         & 90.18         & 44.85        & 42.68        & 41.16        \\
        Oracle (BS)    & 91.06         & 90.66         & 90.38         & 43.32        & 41.46        & 40.18        \\[0.5ex]
        \Xhline{1\arrayrulewidth} \\[-1.5ex]
        SimCLS         & 88.92         & 88.87         & 88.82         & 37.24        & 36.95        & 36.65        \\
        BRIO-Ctr       & 89.03         & 88.93         & 88.85         & \textbf{38.06} & \textbf{37.55} & \textbf{37.14} \\[0.5ex]
        \Xhline{1\arrayrulewidth} \\[-1.5ex]
        BalSum         & \textbf{89.67}  & \textbf{89.60} & \textbf{89.54} & 37.46        & 37.08        & 36.78        \\[0.5ex]
        \Xhline{1pt}\\
    \end{tabular}%
    }
    \vspace{-1\baselineskip}
    \caption{Analysis of re-ranking performance on CNN/DM. \textbf{BS} and \textbf{R} denote BERTScore and the mean ROUGE $F_{1}$ score, respectively. Oracle (R) is ordered by ROUGE scores, while Oracle (BS) is ordered by BERTScore.}   
    \label{tab:tab_analysis}
}
\end{table}

\begin{table}[t]
\small{
    \centering
    \begin{tabular}{lcccc}
    \Xhline{1pt} \\[-1.5ex]
                    & \multicolumn{2}{c}{\textbf{CNNDM}} & \multicolumn{2}{c}{\textbf{XSum}} \\[0.5ex]
    \textbf{Model} & \textbf{$F_1$}      & \textbf{FP}($\%$)     & \textbf{$F_1$}     & \textbf{FP}($\%$)     \\ \Xhline{1\arrayrulewidth} \\[-1.5ex]
    BRIO-Ctr       & 78.50            & 10.96           & \textbf{76.95}  & 10.01           \\
    BalSum         & \textbf{78.84}   & 10.73           & 76.32           & 10.49           \\
    \Xhline{1pt}\\
    \end{tabular}
    \vspace{-1\baselineskip}
    \caption{$F_1$ score and percentage of false positives on all two datasets. The high $F_1$ score indicates how well the ranking model estimates both lexical and semantic quality of all candidate summaries in the pool. \textbf{FP} stands for false positives.}
    \label{tab:F1_and_FP}
}
\end{table}

\paragraph{Ranking Evaluation} Regardless of the number of candidates, an ideal ranking model should yield oracle results considering diverse aspects of summarization. We conduct an experiment to measure the qualities by selecting the top-$k$ summaries after aligning the candidates through different models. As shown in Table \ref{tab:tab_analysis}, we can see that our model shows consistent performance in both evaluation metrics depending on the $k$ (about $\pm0.06$ BERTScore, $\pm0.34$ ROUGE average score). Compared to SimCLS and BRIO-Ctr, the second block in Table \ref{tab:tab_analysis} demonstrates that BalSum captures semantic similarity best while maintaining the intermediate level from the perspective of lexical overlap quality. Moreover, we find that BalSum has the lowest drop ratio of BERTScore ($-1.52\%$) from the perfect ranking ``oracle'' scores. 

We also investigate whether all ranked summaries by models satisfy both lexical and semantic quality. We evaluate models using $F_1$ which measures the cases where the higher-ranked summary has both larger ROUGE and BERTScore than the lower-ranked summary. In addition, we calculate the percentage of false positives. Following Table \ref{tab:F1_and_FP}, while BalSum has worse ($+0.48\%$ FP, $-0.63$ $F_1$) than BRIO-Ctr on XSum, it has better ranking performance ($-0.23\%$ FP, $+0.34$ $F_1$) on CNN/DM. We observe that the decrease of false positives leads to an improvement in $F_1$ score, demonstrating that the result of Table \ref{tab:CNNDM} can be interpreted as reducing semantic mistakes in ranking. As a result, we find that (1) our model is able to learn how to score each summary by balancing the lexical and semantic quality, and (2) the other reason of weak performance on XSum is related to small decline of false positives compared to CNN/DM. 

\paragraph{Case Study on CNN/DM} Table \ref{tab:case_study} presents an intriguing pattern we observed when comparing the results of BRIO-Ctr and BalSum, which demonstrate that our model helps to capture precise details from documents. While BRIO-Ctr contains some irrelevant information in the summaries (shown as \ctext[RGB]{97,189,252}{highlighted text in blue}), BalSum selects the summaries where the last sentence is more consistent with the reference (shown as \ctext[RGB]{255,215,0}{highlighted text in yellow}). Furthermore, despite the comparable ROUGE scores of both models, we note that BalSum's selected summaries consistently have higher BERTScore than those of BRIO-Ctr. 

\section{Conclusion}
In this work, we propose BalSum which aims to evaluate summaries by considering the balance between lexical and semantic quality. To achieve this, we perform a multi-task learning, which aligns summaries according to their lexical overlap qualities and identifies whether they are similar to the document. In addition, to our best knowledge, our method is the first attempt to present a new perspective of false positives (semantic mistakes) in ranking and creating the model to reduce their influence. Our experimental results and fine-grained analyses validate that our model achieves consistent improvements over competitive baselines. 

\section*{Limitations}
% \paragraph{Computation Resources} We set the size of the candidate summary pool to 16, as it is close to the maximum which could fit in a standard 24GB RAM GPU. As shown in Appendix \ref{sec:appendix_num_candi}, even though we demonstrate that our method is robust to the number of candidates, it is not available to train more candidate summaries in our experimental environment. 
\paragraph{Candidate Summaries Dependency} While we mainly investigate a training objective to select the best summary among a set of candidates, we find that our model has been dependent on those obtained from the generation model. Recently, several works have been presented to improve language generation. For example, \citet{narayan-etal-2022-well} and \citet{xu-etal-2022-massive} improve decoding methods to generate diverse outputs. It will be beneficial when applying our method to these approaches. 
\paragraph{One-sentence Summary} Our approach can fail to capture the information from an extremely short summary. Since Table \ref{tab:XSum} shows that our approach has a smaller improvement than CNN/DM, we plan to investigate that our model aims to capture more detailed features from an input text.  

\section*{Acknowledgements}
We thank Soohyeong Kim and anonymous reviewers for valuable feedback and helpful suggestions. This work was supported by the National Research Foundation of Korea(NRF) grant funded by the Korea government(*MSIT) (No.2018R1A5A7059549 , No.2020R1A2C1014037) and supported by Institute of Information \& communications Technology Planning \& Evaluation (IITP) grant funded by the Korea government(*MSIT) (No.2020-0-01373). *Ministry of Science and ICT

\bibliography{anthology,custom}
\bibliographystyle{acl_natbib}

\appendix
\section{Distribution of $z$ on XSum}
\label{sec:appendix_xsum}
\begin{figure}[!ht]
    \centering
    \includegraphics[width=1.0\linewidth]{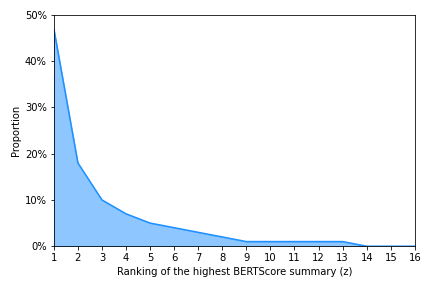}
    \caption{Distribution of z(\%) for a base PEGASUS model on XSum. Because a PEGASUS model generates a pool of 16 diverse beam search candidates, the X-axis ranges from 1 to 16. The Y-axis represents the proportion of $z$ in the test set.}
    \label{fig:appendix_xsum}
\end{figure}

The result in Fig. \ref{fig:appendix_xsum} shows that there is a majority (53\%) of cases where $z>1$.

\section{Evaluation Metrics}
\label{sec:appendix_eval} 
We examine our model with two evaluation metrics.
\begin{itemize}
    \item \textbf{ROUGE} \citep{lin-2004-rouge} is a widely used metric for summarization evaluation. We use the standard ROUGE Perl package\footnote{\url{https://github.com/summanlp/evaluation/tree/master/ROUGE-RELEASE-1.5.5}} for evluation. 
    \item \textbf{BERTScore} \citep{2019-BERTScore} is a semantic similarity metric for multiple tasks. We use the public \textit{bert-score} package\footnote{\url{https://github.com/Tiiiger/bert_score}} shared by the authors. 
\end{itemize}

\section{Datasets Statistics}
\label{sec:appendix_dataset}
\begin{table}[!h]
    \centering
    \small{
    \begin{tabular}{llll}
    \Xhline{1pt} \\[-1.5ex]
        \textbf{Dataset} & \textbf{Train} & \textbf{Valid} & \textbf{Test} \\[0.5ex]
        \Xhline{1\arrayrulewidth} \\[-1.5ex]
        CNN/DM & 287,227 & 13,368 & 11,490 \\
        XSum & 204,045 & 11,332 & 11,334 \\[0.5ex]
    \Xhline{1pt} \\[-1.5ex]
    \end{tabular}
    \caption{Statistics of two datasets}
    }
    \label{tab:dataset}
\end{table}

\section{Implementation Details}
\label{sec:appendix_details}
\paragraph{Model} We implement our model based on Huggingface Transformers library \citep{wolf-etal-2020-transformers}. We use the pre-trained RoBERTa with ‘roberta-base’ version, containing around 125M parameters. Our experiments are conducted on a single NVIDIA RTX 3090 GPU with 24GB memory. 

\paragraph{Decoding Setttings} We use the diverse beam search algorithm \citep{VijayakumarCSSL16} to decode summaries. We generate candidate summaries from 16 diversity groups with 16 beams. On CNN/DM and XSum, we use the pre-trained BART\footnote{The checkpoint is ``facebook/bart-large-cnn'', containing around 400M parameters.} and PEGASUS\footnote{The checkpoint is ``google/pegasus-xsum'', containing around 568M parameters.} models as the generation model. 

\paragraph{Training Settings} We train our models for 5 epochs using an Adafactor optimizer \citep{pmlr-v80-shazeer18a}. The batch size is 4 and the learning rate is 2e-3. During training, we randomly select 4 negative samples for each input document. We evaluate the model every 1000 steps on the validation set.

\newpage
\section{Effect of Model Architecture}
\label{sec:appendix_WeightAvg}
We train BalSum with different model architectures and evaluate them on CNN/DM test set. For a fair comparison, we use only ranking loss in Eq. \ref{eq5}. Table \ref{tab:Weighted_Sum} shows that taking the weighted sum of scores in Eq. \ref{eq4} leads to better performance than others.

\begin{table}[!h]
\small{
    \centering
    \begin{tabular}{llll}
    \Xhline{1pt} \\[-1.5ex]
    \textbf{Model} & \textbf{R-1}   & \textbf{R-2}   & \textbf{R-L}   \\[0.5ex]
    \Xhline{1\arrayrulewidth} \\[-1.5ex]
    \textit{[CLS]} & 45.40          & 21.18          & 42.36          \\
    \textit{Avg.}  & 46.59          & \textbf{22.40} & 43.47          \\
    Ours           & \textbf{46.64} & 22.38          & \textbf{43.52} \\[0.5ex]
    \Xhline{1pt} \\[-1.5ex]
    \end{tabular}
    \caption{Ablation studies of different model architectures on CNN/DM. \textbf{R-1/2/L} denotes ROUGE-1/2/L. \textit{[CLS]}: using the first [CLS] token. \textit{Avg.}: averaging all scores in Eq. \ref{eq3}.}
    \label{tab:Weighted_Sum}
}
\end{table}

\section{Identical Candidates Scores}
As shown in Table \ref{tab:identi}, we note cases that have at least two identical R-avg on CNN/DM and XSum are a majority. Since we count after removing the same summaries in the pool, we ensure that it is the number of summaries with different content but the same R-avg score. 
\label{sec:appendix_identical_score}
\begin{table}[!h]
\small{
    \resizebox{.5\textwidth}{!}{%
    \centering
    \begin{tabular}{ccP{.1\textwidth}P{0.2\textwidth}}
    \Xhline{1pt} \\[-1.5ex]
\textbf{Dataset} & \textbf{Decoding methods} & \textbf{\# Summary candidates} & \textbf{\# of pools with at least two same R-avg (\%)} \\[0.5ex]
    \Xhline{1\arrayrulewidth} \\[-1.5ex]
    CNN/DM & Diverse beam search & 16 & 46.09 \\[0.5ex]
    Xsum & Diverse beam search & 16 & 73.01 \\[0.5ex]
    \Xhline{1pt} \\[-1.5ex]
\end{tabular}%
}
}
\caption{Number of pools with at least two same R-avg (\%). A pool consists of 16 diverse beam search candidates generated on different datasets (CNN/DM, XSum) with different base models (PEGASUS, BART). R-avg is the average of ROUGE-1/2/L scores. }
    \label{tab:identi}
\end{table}

\begin{table*}[!t]
    \resizebox{\textwidth}{!}{%
    \centering
    \begin{tabular}{p{.1\textwidth} p{.07\textwidth} p{.05\textwidth} p{.8\textwidth}}
    \Xhline{1pt} 
    \textbf{System} & \textbf{R-avg} & \textbf{BS} & \textbf{Summary} \\
    \Xhline{1\arrayrulewidth} 
        % \small{Article} & $-$ & $-$ & \small{HI~} \\
        % \Xhline{1\arrayrulewidth}
        \small{Reference} & $-$ & $-$ & \small{Didier Drogba played first Chelsea game after joining on free from Galatasaray. Ivory Coast striker was second half substitute for Diego Costa in 3-0 defeat by Werder Bremen. John Terry handed him captaincy later in game, \textbf{but 36-year-old failed to trouble German side in front of goal.}} \\
        \Xhline{1\arrayrulewidth}
        \small{Diverse beam \#1} & \small{$30.72$} & \small{$87.50$} & \small{Ivory Coast striker made his second return to the club. Drogba was a half-time substitute in the 3-0 defeat at the Weserstadion. The 36-year-old was replaced by Diego Costa at half- time. \textcolor{ForestGreen}{Dobar was the first player on the pitch after John Terry left.}} \\
        \hdashline
        \small{Diverse beam \#2} & \small{$28.39$} & \small{$91.17$} & \small{Didier Drogba made his second Chelsea debut in pre-season friendly at Werder Bremen. The 36-year-old was a half-time substitute as Chelsea lost 3-0. Drogbba was captain after John Terry left the pitch in the second half. \textcolor{red}{The Ivorian striker missed a penalty and failed to make an impact on the game.}} \\
    \Xhline{1pt} 
    \end{tabular}%
}
    \caption{False positive examples from fine-tuned BART model on CNN/DM. \textbf{R-avg} is the average of ROUGE-1/2/L scores. \textbf{BS} denotes BERTScore. The related sentences in the reference are in \textbf{bold}.}
    \label{tab:appendixC}
\end{table*}

\section{Examples for False Positive}
\label{sec:appendix_example}
Table. \ref{tab:appendixC} shows that \#2 has 2.33 R-avg lower than \#1, but 3.67 BERTScore higher. Also, when evaluated qualitatively, it can be seen that \#2 is closer to the gold summary. While \textcolor{ForestGreen}{the sentence in green} is discarded, \textcolor{red}{the sentence in red} is included in the reference summary. 

\newpage
\section{Negative Size and Scale Factors}
\label{sec:appendix_hyper}
\begin{figure}[!h]
    \centering
    \includegraphics[width=1.0\linewidth]{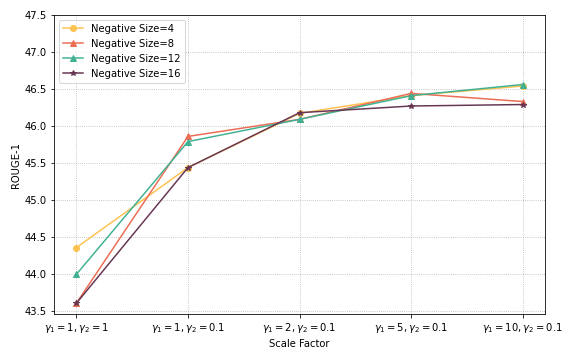}
    \caption{ROUGE-1 on CNN/DM w.r.t scale factors and $N$ negative samples at inference time, with $N \in \{ 4, 8, 12, 16 \}$.}
    \label{fig:fig_appendixG}
\end{figure}

We have tuned the scale factor $\gamma_1$ of ranking loss and $\gamma_2$ of contrastive loss in Eq. \ref{eq8} with different sizes of negative samples. As shown in Fig. \ref{fig:fig_appendixG}, suitable scale factors ($\gamma_1=10, \gamma_2=0.1$) can improve more than others. Though $size=4$ and $size=12$ showed similar performance, we set the negative size to 4 due to memory efficiency. 

\section{Number of Candidate Summaries}
\label{sec:appendix_num_candi}
\begin{figure}[!h]
    \centering
    \includegraphics[width=1.0\linewidth]{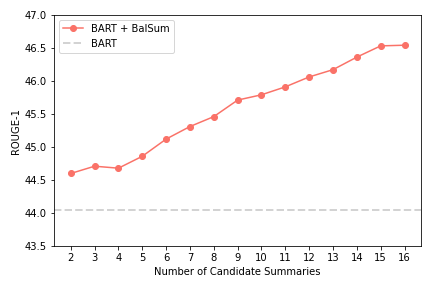}
    \caption{ROUGE-1 with different numbers of candidate summaries on CNN/DM. The gray dashed line denotes the performance of a base model (BART).}
    \label{fig:fig_num_candi}
\end{figure}

We set the size of the candidate summary pool to 16, as it is close to the maximum which could fit in a standard 24GB RAM GPU. Fig. \ref{fig:fig_num_candi} reports that our method is robust to the number of candidates.

\definecolor{Gray}{gray}{0.9}
\begin{table*}[!h]
    \resizebox{\textwidth}{!}{%
    \centering
    \rowcolors{1}{}{Gray}
    \begin{tabular}{p{.1\textwidth} p{.05\textwidth} p{.05\textwidth} p{.05\textwidth} p{.05\textwidth} p{.8\textwidth}}
\hline
\textbf{System} & \textbf{R-1} & \textbf{R-2} & \textbf{R-L} & \textbf{BS} & \textbf{Summary}         \\ 
\hline
Reference & -           & -           & -           & -           & \small{arsene wenger will have chat with theo walcott ahead of arsenal clash. walcott was substituted after 55 minutes of england's draw with italy. arsenal boss is wenger is concerned by the winger's confidence. \textbf{the gunners take on liverpool at the emirates stadium on saturday.}} \\
BRIO-Ctr & \small{60.61} & \small{41.24}       & \small{46.46} & \small{89.93} & \small{theo walcott played just 55 minutes of england's 1-1 draw with italy. arsene wenger says he is concerned by the winger's confidence. the arsenal manager will speak with walcott ahead of liverpool clash. \ctext[RGB]{97,189,252}{walcott could start against liverpool on saturday with alex oxlade-chamberlain out and danny welbeck a doubt.}} \\ 
BalSum & \small{61.54} & \small{38.20} & \small{41.76} & \small{92.36} &  \small{arsenal winger theo walcott struggled for england against italy. arsene wenger says he is concerned by the winger's confidence. walcott was replaced after 55 minutes of england's 1-1 draw in turin. \ctext[RGB]{255,215,0}{the gunners face liverpool on saturday in a top-four clash.}} \\ \hline
Reference & -           & -           & -           & -           &  \small{experts have voiced concerns over diy brain stimulation kits for children. for a few hundred dollars, one can be purchased online from various sites. it promises to help children with math homework and claims to help adhd. professor colleen loo from the black dog institute strongly believes that the equipment poses a danger to amateurs and children. \textbf{the equipment is currently being used to treat people with speech impediments but is still very much in trial stages.}} \\
BRIO-Ctr & \small{40.0 } & \small{16.26} & \small{19.20} & \small{87.11} & \small{for a few hundred dollars, you can purchase a brain stimulation kit online. experts have voiced concerns over the potential side effects. the kits are being sold online for as little as \$ 55 us. \ctext[RGB]{97,189,252}{one site even advertises how to make your own electrodes using a household sponge.}} \\
BalSum & \small{36.92} & \small{17.19} & \small{27.69} & \small{89.90} & \small{parents are buying diy brain stimulation kits for their children. the kits are being sold online for as little as \$ 55 us. experts are concerned about the potential side effects of the equipment. the devices are used to improve speaking in those with speech problems. \ctext[RGB]{255,215,0}{the equipment is still relatively new and experimental.}} \\ \hline
Reference & -           & -           & -           & -           & \small{ross barkley has been repeatedly linked with a move to manchester city. former city star gareth barry says his everton team-mate is too young. \textbf{the toffees face manchester united in the premier league on sunday.}} \\
BRIO-Ctr & \small{47.19} & \small{27.59} & \small{29.21} & \small{88.85} & \small{everton team-mate gareth barry has advised ross barkley against moving to manchester city. the 21-year-old has been linked with a move away from goodison park. barry believes it is too early for the youngster to decide on his future. \ctext[RGB]{97,189,252}{the veteran midfielder spent four seasons at the etihad before joining everton.}} \\ 
BalSum & \small{46.34} & \small{25.0} & \small{34.15} & \small{91.16} & \small{gareth barry has advised ross barkley against moving to manchester city. the everton midfielder believes it is too early for the 21-year-old to decide on his future. barry spent four seasons at the etihad before arriving on merseyside. \ctext[RGB]{255,215,0}{the toffees face manchester united on sunday.}} \\ \hline
Reference & -           & -           & -           & -           & \small{local councils are urged to draw up maps of the residents who are at risk. \textbf{essex and gloucestershire have already made ‘loneliness maps’ experts warn that being lonely can lead to serious health problems.}} \\
BRIO-Ctr & \small{50.57} & \small{28.24} & \small{29.89} & \small{90.30} & \small{two county councils have already implemented ‘loneliness maps’ to target ‘danger zones’ being lonely can lead to health problems including dementia and high blood pressure. campaigners say councils should draw up maps of the places where pensioners are most at risk. \ctext[RGB]{97,189,252}{study by university of kent and campaign to end loneliness recommends maps.}} \\
BalSum & \small{50.0} & \small{27.91} & \small{43.18} & \small{91.28} & \small{campaigners say councils should draw up maps of places where pensioners and others are most likely to suffer from social isolation. two county councils, essex and gloucestershire, have already implemented the maps. they allow them to target ‘danger zones’ of loneliness. \ctext[RGB]{255,215,0}{being lonely can lead to health problems including dementia and high blood pressure.}} \\ \hline
Reference & -           & -           & -           & -           & \small{the gruesome vision was captured in australia and uploaded last week. the lizard swings its neck back and forth in a bid to swallow the rabbit. \textbf{goannas can unhinge their lower jaws allowing them to swallow large prey.}} \\
BRIO-Ctr & \small{51.16} & \small{23.81} & \small{27.91} & \small{88.75} & \small{two-metre long reptile is filmed balancing on top of a power pole to swallow rabbit. the lizard swings its neck back and forth as it battles to swallow its catch. \ctext[RGB]{97,189,252}{it finishes the feat in under a minute, and the video was uploaded to youtube last week.}} \\
BalSum & \small{46.91} & \small{20.25} & \small{34.57} & \small{90.72} & \small{two-metre long lizard filmed battling to swallow rabbit in under one minute. video shows lizard balance at the top of a power pole while swallowing its prey. \ctext[RGB]{255,215,0}{goannas can unhinge their lower jaws when feeding, allowing them to eat over-sized prey.}} \\ \hline
\end{tabular}%
}
    \caption{\textbf{Case Study} on CNN/DM. R-1/2/L are the ROUGE-1/2/L $F_{1}$ scores. \textbf{BS} denotes BERTScore. The related sentences in the reference are in \textbf{bold}.}
    \label{tab:case_study}
\end{table*}
\end{document}